\documentclass[runningheads]{llncs}
\usepackage{graphicx}
\usepackage{comment}
\usepackage{amsmath,amssymb} 
\usepackage{color}

\usepackage{amsmath}
\usepackage{amssymb}
\usepackage{multirow}
\usepackage{booktabs}
\usepackage{cite}

\usepackage{algorithm}
\usepackage[noend]{algpseudocode}

\usepackage[font=small,labelfont=bf]{caption}

\newcommand{\eqnref}[1]{{Eq.\ \eqref{eq:#1}}}
\newcommand{\secref}[1]{{Section \ref{sec:#1}}}
\newcommand{\figref}[1]{Figure\,\ref{fig:#1}}
\newcommand{\tabref}[1]{Table~\ref{tab:#1}}

\def\eg{\emph{e.g}}
\def\etal{\textit{et al. }}

\begin{document}
\pagestyle{headings}
\mainmatter
\def\ECCVSubNumber{1340}  

\title{Procedure Planning in Instructional Videos} 

\titlerunning{Procedure Planning in Instructional Videos}
%

\author{Chien-Yi Chang\inst{1} \and
De-An Huang\inst{1} \and
Danfei Xu\inst{1} \and
Ehsan Adeli\inst{1} \and
Li Fei-Fei\inst{1} \and
Juan Carlos Niebles\inst{1}}

\authorrunning{Chang et al.}
%
\institute{Stanford University, Stanford CA 94305, USA}

\maketitle


\begin{abstract}

In this paper, we study the problem of procedure planning in instructional videos, which can be seen as a step towards enabling autonomous agents to plan for complex tasks in everyday settings such as cooking. Given the current visual observation of the world and a visual goal, we ask the question ``What actions need to be taken in order to achieve the goal?". The key technical challenge is to learn structured and plannable state and action spaces directly from unstructured videos. We address this challenge by proposing Dual Dynamics Networks (DDN), a framework that explicitly leverages the structured priors imposed by the conjugate relationships between states and actions in a learned plannable latent space. We evaluate our method on real-world instructional videos. Our experiments show that DDN learns plannable representations that lead to better planning performance compared to existing planning approaches and neural network policies.

\end{abstract}
\section{Introduction}

\begin{figure}
  \centering

\includegraphics[width=1.0\linewidth]{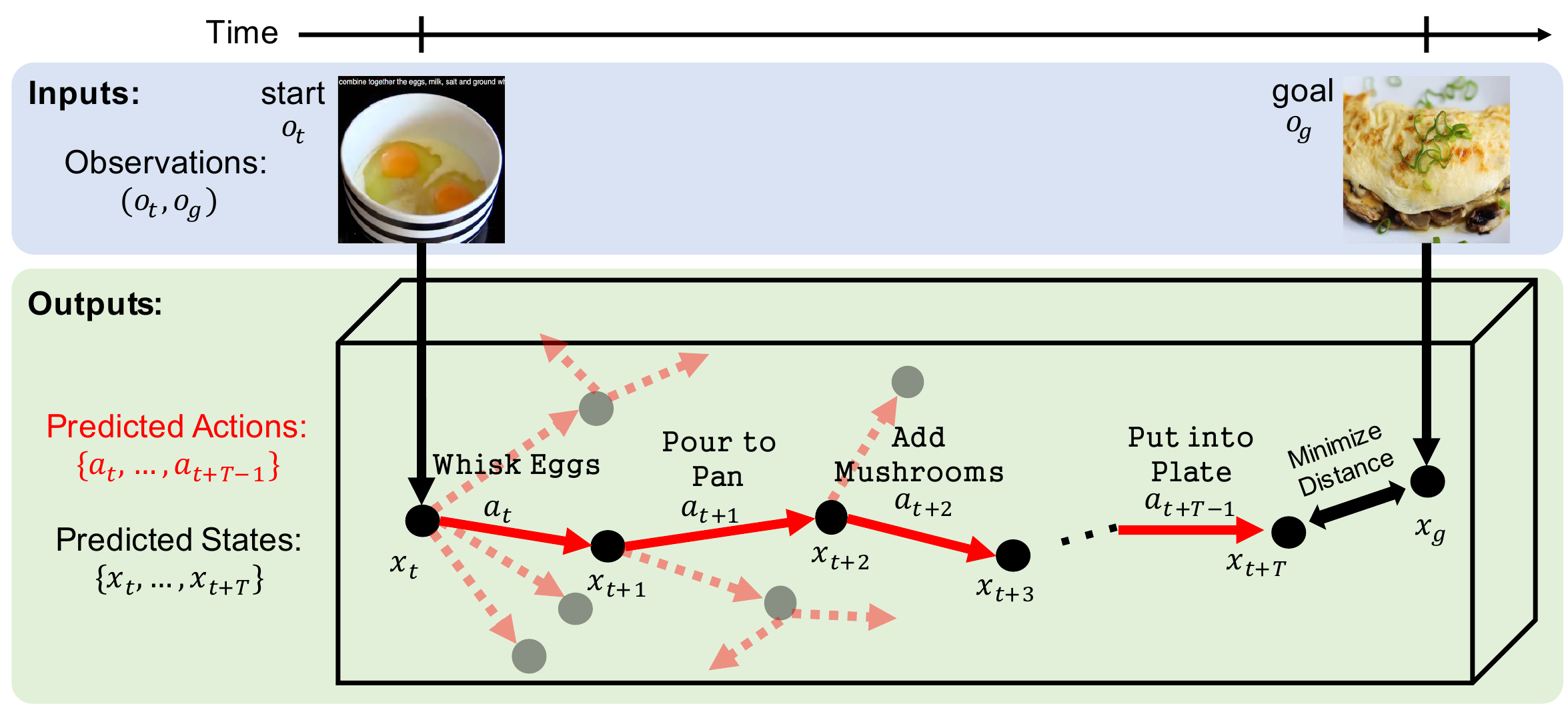}
  \vspace{-6mm}

    \caption{We study the problem of \emph{procedure planning} in instructional videos. The goal is to generate a sequence of actions towards a desired goal such as making a mushroom omelette. Given a start observation $o_t$ and a visual goal $o_g$, our model plans a sequence of actions $\{a_{i}\}$ (red arrows) that can bring the start towards the goal. In addition, our model also predicts a sequence of intermediate states $\{x_{i}\}$ (black dots).}
  \label{fig:fig1}
    \vspace{-3mm}

\end{figure}

What does it take for an autonomous agent to perform complex tasks in everyday settings, such as cooking in a kitchen? The most crucial ability is to know what actions should be taken in order to achieve the goal. In other words, the agent needs to make goal-conditioned decisions sequentially based on its perception of the environment. This sequential decision making process is often referred to as \emph{planning} in the robotics literature. While recent works have shown great promise in learning to plan for simple tasks in structured environments such as pushing objects or stacking blocks on a table~\cite{finn2017deep,finn2016deep,srinivas2018universal}, it is unclear how these approaches will scale to visually complex, unstructured environments as seen in instructional videos ~\cite{zhukov2019cross,zhou2018towards} (recordings of human performing everyday tasks).

In this paper, we study procedure planning in instructional videos, which can be seen as a step towards enabling autonomous agents to plan for complex tasks in everyday settings. Given the current visual observation of the world and a visual goal, we ask the question ``What actions need to be taken in order to achieve the goal?". 
As illustrated in \figref{fig1}, we define the \emph{procedure planning} problem as: given a current visual observation $o_t$ and a visual goal $o_g$ that indicates the desired final configuration, the model should plan a sequence of high-level actions $ \{a_t, \cdots, a_{t+T-1}\}$ that can bring the underlying state of $o_t$ to that of $o_g$. Here, $T$ is the horizon of planning, which defines how many steps of task-level actions we allow the model to take.

The key technical challenge of the proposed procedure planning problem is how to learn structured and plannable state and action spaces directly from unstructured real videos. Since instructional videos are visually complex and the tasks are often described at high levels of abstractions, one can imagine indefinitely growing semantic state and action spaces from the visually complex scenes and high-level descriptions, which prevents the application of classical symbolic planning approaches~\cite{ghallab2004automated, konidaris2018skills} as they require a given set of predicates for a well-defined state space.

We address this challenge by explicitly leveraging the conjugate relationships between states and actions to impose structured priors on the learned latent representations. 
Our key insight is that in addition to modeling the action as a transformation between states, we can also treat the state as the precondition for the next action because the state contains the history of previous actions. Following this intuition, we jointly train two modules: a forward dynamics model that captures the transitional probabilities between states and a conjugate dynamics model that utilizes the conjugate constraints for better optimization. We call our proposed method Dual Dynamics Networks (DDN), a neural network framework that learns plannable representations from instructional videos and performs procedure planning in the learned latent space. 

We evaluate our approach on real-world instructional videos~\cite{zhukov2019cross} and show that the learned latent representations and our DDN formulation are effective for planning under this setting. Our approach significantly outperforms existing planning baselines for procedure planning. We further show that our model is able to generalize to variations of the start and goal observations. In addition, we show that our approach can be applied to a related problem called walkthrough planning~\cite{kurutach2018learning} and outperforms existing methods.

Our contributions are three-fold. 
First, we introduce the problem of procedure planning in instructional videos, which can be seen as a step towards enabling autonomous agents to plan for complex tasks in everyday settings such as cooking in a kitchen. 
Second, we propose Dual Dynamics Network (DDN), a framework that explicitly leverages structured priors imposed by the conjugate relationships between states and actions for latent space planning. 
Third, we evaluate our approach extensively and show that it significantly outperforms existing planning and video understanding methods for the proposed problem of procedure planning. 
\section{Related Work}

\noindent\textbf{Task Planning.} 
\emph{Procedure planning} is closely related to task planning widely studied in classical AI and robotics. Task planning is the problem of finding a sequence of task-level actions to reach the desired goal state from the current state. In the task planning literature, most studies rely on a pre-defined planning problem domain for the task~\cite{mcdermott1998pddl,ghallab2004automated,konidaris2018skills}. Our work diverges from those because our proposed model can perform task-level, long-horizon planning in the visual and semantic space without requiring a hand-defined symbolic planning domain.

\noindent\textbf{Planning from Pixels.} 
Recent works have shown that deep networks can learn to plan directly from pixel observations in domains such as table-top manipulation~\cite{kurutach2018learning, srinivas2018universal, sermanet2018time}, navigation in VizDoom~\cite{pathak2017curiosity}, and locomotion in joint space~\cite{ehsani2018let}. 
However, learning to plan from unstructured high-dimensional observations is still challenging~\cite{finn2017deep,hafner2018learning}, especially for long-horizon, complex tasks that we want to address in procedure planning. 
A closely related method is Universal Planning Networks (UPN)~\cite{srinivas2018universal}, which uses a gradient descent planner to learn representations from expert demonstrations. However, it assumes the action space to be differentiable. Alternatively, one can also learn the forward dynamics by optimizing the data log-likelihood from the actions~\cite{hafner2018learning}. We use a similar formulation and further propose the conjugate dynamics model to expedite the latent space learning. Without using explicit action supervision, causal InfoGAN~\cite{kurutach2018learning} extracts state representations by learning salient features that describe the causal structure of toy data. In contrast to~\cite{kurutach2018learning}, our model operates directly on real-world videos and handle the semantics of actions with sequential learning.

\noindent\textbf{Understanding Instructional Videos.} 
There has been a growing interest in analyzing instructional videos~\cite{tang2019coin, zhou2018towards,zhukov2019cross, kuehne2014language, miech2019howto100m} by studying a variety of challenging tasks. Some of the tasks ask the question ``What is happening?", such as action recognition and temporal action segmentation~\cite{huang2016connectionist, richard2018neuralnetwork,chang2019d3tw}, state understanding~\cite{alayrac2017joint}, video summarization/captioning~\cite{sun2019videobert,zhou2018towards, sener2019zero}, retrieval~\cite{sun2019videobert} etc. The others ask the question ``What is going to happen?", such as early action recognition~\cite{wang2019progressive, furnari2019would}, action label prediction~\cite{abu2018will,rhinehart2017first,zeng2017visual,sener2019zero, mehrasa2019variational, ke2019time, farha2019uncertainty, furnari2019would}, video prediction~\cite{lan2014hierarchical,ranzato2014video,oh2015action, lee2018stochastic,lu2017flexible,vondrick2016anticipating}, etc. However, due to the large uncertainty in human activities, the correct answer to this question is often not unique. In this paper, we take a different angle and instead ask the question: ``What actions need to be taken in order to achieve a given visual goal?" Rather than requiring the model to predict potentially unbounded future actions conditioned only on history, we ask the model to infer future actions conditioned on both history and goal. Such goal-conditioned formulation immediately resolves the ambiguity when the history does not encode adequate information to bound future actions.

\section{Method}

We are interested in planning in real-world instructional videos. 
The key technical challenge is how to learn structured and plannable state and action spaces directly from unstructured real videos. We take a latent space approach by learning plannable representations of the visual observations and actions, along with the forward and conjugate dynamics models in the latent space. We will first define the procedure planning problem setup and how to address it using a latent space planning approach. We will then discuss how we learn the latent space and leverage the conjugate relationships between states and actions to avoid trivial solutions to our optimization. Finally, we will present the algorithms for procedure planning and walkthrough planning~\cite{kurutach2018learning} in the learned plannable space.

\subsection{Problem Formulation}

As illustrated in \figref{fig1}, given a current visual observation $o_t$ and a visual goal $o_g$ that indicates the desired final configuration, we aim to plan a sequence of actions $\pi = \{a_t, \cdots, a_{t+T-1}\}$ that can bring the underlying state of $o_t$ to that of $o_g$. $T$ is the horizon of planning, which defines how many steps of task-level actions we allow the model to take. 
We formulate the problem as latent space planning~\cite{kurutach2018learning,srinivas2018universal}. Concretely, we learn a model that can plan in some latent space spanned by the mapping functions $f$ and $g$ that encode the visual observation $o$ and action $a$ to a semantic state $f(o) = x$ (\eg from the observed frames to \verb|cooked eggs with mushrooms|) and a latent action state $g(a) = \bar a$ respectively.

Inspired by the classical Markovian Decision Process (MDP)~\cite{bellman1957markovian}, we assume that in this latent space there exists a forward dynamics $\mathcal{T}(x_{t+1}|x_t, \bar a_t)$ that predicts the future state $x_{t+1}$  given the current state $x_t$ and the applicable action $\bar a_t$. Using the forward dynamics, we can perform sampling based planning~\cite{ghallab2004automated} by applying different actions and search for the desired goal state $x_g$. Specifically, given $f(\cdot)$, $g(\cdot)$ and $\mathcal{T}$, we can find a plan $\pi = \{a_t, \cdots, a_{t+T-1}\}$ by (i) mapping from the visual space to the latent space $x_t = f(o_t)$, $x_g = f(o_g)$ and (ii) search in the latent space using $\mathcal{T}(x_{t+1}|x_t, \bar a_t)$ to find the sequence of actions that can bring $x_t$ to $x_g$. We will discuss details of this procedure in \secref{plan}.

\subsection{Learning Plannable Representations}
\label{sec:learn}

In this section, we discuss how to learn the embedding functions $f(\cdot)$, $g(\cdot)$, and forward dynamics $\mathcal{T}$ from data. 
One possible approach is to directly optimize $f(\cdot)$ with some surrogate loss function such as mutual information~\cite{kurutach2018learning}, without explicitly modeling $\mathcal{T}(\cdot|x, \bar a)$ and $g(\cdot)$. 
This approach is limited because it assumes a strong correspondence between the visual observation $o$ and the semantic state $x = f(o)$. 
In real-world videos, however, a small change in visual space $\Delta o$ can induce a large variation in the semantic space $\Delta x$ and vice versa. An alternative approach is to formulate a differentiable objective jointly with $f(\cdot)$, $g(\cdot)$, and $\mathcal{T}(\cdot|x, \bar a)$~\cite{srinivas2018universal}. This approach also falls short when applied to procedure planning, because it often requires the action space to be continuous and differentiable, while we have an unstructured and discrete action space in real-world instructional videos.

\begin{figure}[t]
  \centering
  \includegraphics[width=1.0\linewidth]{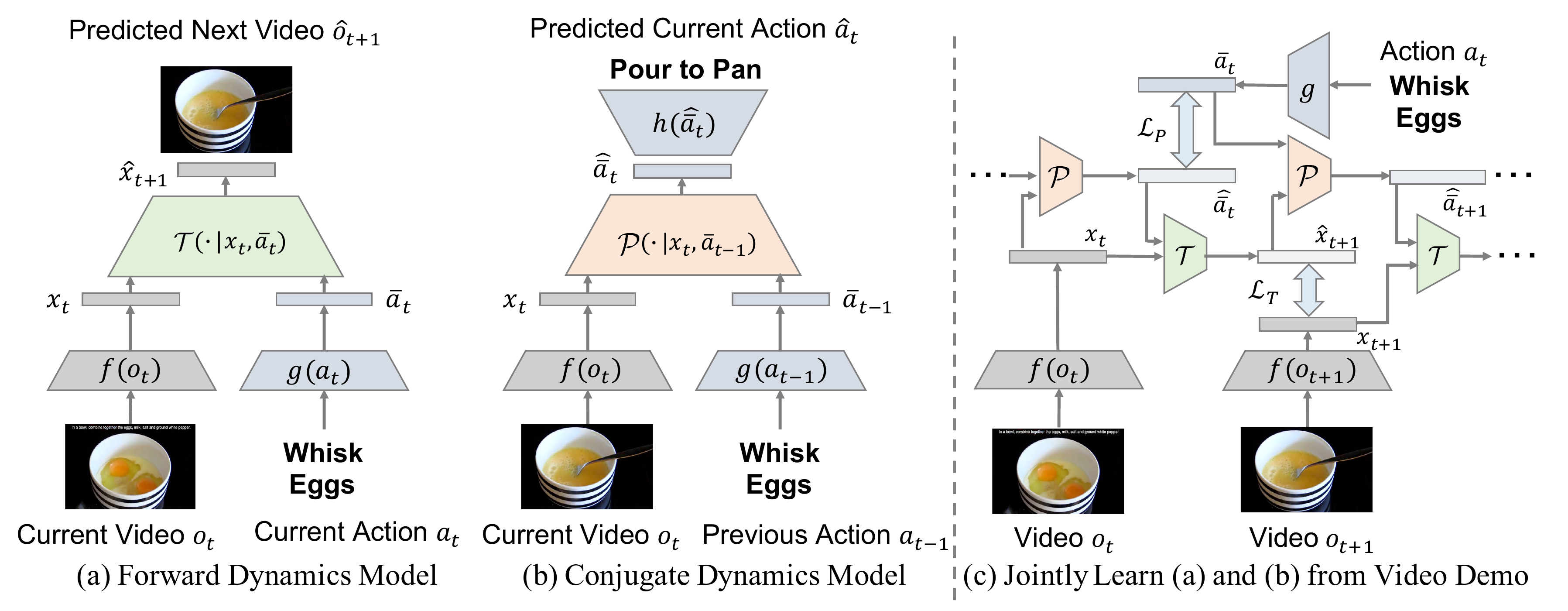}
  \vspace{-6mm}
  \caption{(a) Our forward dynamics model $\mathcal{T}$ predicts the next state based on the current state and action. (b) We learn the conjugate dynamics model $\mathcal{P}$ jointly with $\mathcal{T}$ to restrict the possible state mapping $f$ and action embedding $g$. (c) At training time, the forward dynamics model $\mathcal{T}$ takes as inputs the current state $x_t$ and predicted current action $\hat{\bar{a}}_t$ to predict the next state $\hat x_{t+1}$, and the conjugate dynamics model $\mathcal{P}$ takes as inputs the predicted next state $\hat x_{t+1}$ and the current action $\bar a_t$ to predict the next applicable action $\hat{\bar{a}}_{t+1}$. }
  \vspace{-3mm}
  \label{fig:system}
\end{figure}

\vspace{1mm}
\noindent\textbf{Forward Dynamics.} 
To address the above challenges, we propose to learn $f(\cdot)$, $g(\cdot)$, and $\mathcal{T}(\cdot|x, \bar a)$, jointly with a conjugate dynamics model $\mathcal{P}$ which provides further constraints on the learned latent spaces. 
We consider training data $\mathcal{D} = \{(o_t^i, a_t^i, o_{t+1}^i)\}_{i=1}^{n}$, which corresponds to the triplets of current visual observation, the action to be taken, and the next visual observation. The triplets can be curated from existing instructional video datasets and the details will be discussed in Section~\ref{sec:experiments}. The learning problem is to find $\mathcal{T}$ that minimizes:
\begin{align}
\label{eq:t_loss}
\small
\begin{split}
    \mathcal{L}_{T}(\mathcal{T},f, g; \mathcal{D})& = - \frac{1}{n} \sum_{i=1}^{n} \log \Pr(x_{t+1}^{i}|x_t^i, \bar a_t^i; \mathcal{T}) \\
 &= -\frac{1}{n} \sum_{i=1}^{n}  \log \Pr(f(o_{t+1}^{i})|f(o_t^i), g(a_{t}^i); \mathcal{T}).
\end{split}
\end{align}
The architectures are shown in \figref{system}(a).

\vspace{1mm}
\noindent\textbf{Conjugate Dynamics.} There exists a large number of $\mathcal{T}(\cdot|x, \bar a)$ that can minimize $\mathcal{L}_{T}$ and our model can easily overfit to the training data in $\mathcal{D}$. In this case, it is hard for the learned model to generalize to unseen visual observations. It is thus crucial to impose structured priors on the model. Our insight is to leverage the conjugate relationship between states and actions~\cite{hayes2016autonomously} to provide further constraints on $f(\cdot)$ and $g(\cdot)$ and improve the optimization of $\mathcal{L}_{T}$. Please note that this is also how our proposed method diverges from the classical MDP setting and previous works~\cite{srinivas2018universal,kurutach2018learning}.
We can treat the loss in \eqnref{t_loss} as leveraging the standard relationship between states and actions to learn $f(\cdot)$: Given the current semantic state $x_t$, applying a latent action $\bar a_t$ would bring it to a new state $x_{t+1}$. $\mathcal{L}_{T}$ is encouraging $f(\cdot)$ to fit $x_{t+1} = f(o_{t+1})$ and to be consistent with the prediction of $\mathcal{T}(\cdot|x, \bar a)$. On the other hand, the conjugate relationship between states and actions also implies the following: Given the previous action $\bar a_{t-1}$, the current state $x_{t}$ constraints the possible  actions $\bar a_{t}$ because $x_{t}$ needs to satisfy the precondition of $\bar a_{t}$. For example, if $\bar a_{t-1}$ representing \emph{`pour eggs to pan'} is followed by $\bar a_{t}$ representing \emph{`cook it'}, then the state $x_{t}$ in between must satisfy the precondition of $\bar a_{t}$, that is: to cook the eggs, the eggs should be in the pan.
Based on this intuition, we further propose to learn a conjugate dynamics model $\mathcal{P}(\bar a_t|x_t, \bar a_{t-1})$, which is illustrated in \figref{system}(b). The conjugate dynamics model $\mathcal{P}$ takes as inputs the current state and the previous action to predict the current applicable action. Formally, the learning problem is to find $\mathcal{P}$ that minimizes the following function:
\begin{align}
\small
\label{eq:p_loss}
\begin{split}
    \mathcal{L}_{P}(\mathcal{P},f, g, h; \mathcal{D}) 
    &= - \frac{1}{n} \sum_{i=1}^{n} \Big[ \log \Pr(\bar a_{t}^{i}|x_t^i, \bar a_{t-1}^i; \mathcal{P}) + \phi(h(g(a_t^i)), a_t^i) \Big]\\
& =-\frac{1}{n} \sum_{i=1}^{n} \Big[ \log \Pr \left ( g(a_{t}^{i})|f(o_t^i), g(a_{t-1}^i); \mathcal{P} \right ) 
+ \phi(h(g(a_t^i)), a_t^i)  \Big].
\end{split}
\end{align}
where the mapping functions $f$ and $g$ have shared parameters across $\mathcal{T}$ and $\mathcal{P}$. $h$ is an inverse mapping function that decodes the action $a$ from the latent action representation $\bar a$. $\phi(\cdot, \cdot)$ measures the distance in the action space. 

Now we have discussed all the components for Dual Dynamics Network (DDN), a framework for latent space task planning in instructional videos. In our framework, we encode the visual observations and the actions into latent space states and actions with $f(\cdot)$ and $g(\cdot)$, and jointly learn from video demonstrations a forward dynamics model $\mathcal{T}$ that captures the transitional probabilities and a conjugate dynamics model $\mathcal{P}$ that utilizes the conjugate relationship between states and actions by minimizing the following combined loss function:
\begin{align}
\small
\label{eq:combined_loss}
\begin{split}
\mathcal{L}(\mathcal{T}, \mathcal{P}, &f, g, g^{-1}; \mathcal{D}) = \alpha \cdot \mathcal{L}_{T} + \mathcal{L}_{P}.
\end{split}
\end{align}
$\alpha$ is a weighting coefficient for the combined loss.  
The learning process is illustrated in Figure~\ref{fig:system}(c). 
During training, the forward dynamics model $\mathcal{T}$ takes the current state $x_t$ and predicted current action $\hat{\bar{a}}_t$ to predict the next state $\hat x_{t+1}$, and the conjugate dynamics model $\mathcal{P}$ takes the predicted next state $\hat x_{t+1}$ and the current action $\bar a_t$ to predict the next applicable action $\hat{\bar{a}}_{t+1}$.

\subsection{Planning in Latent Space}
\label{sec:plan}

\begin{figure}[t]
\centering
\begin{minipage}{.49\textwidth}
  \centering
  \includegraphics[width=0.9\linewidth]{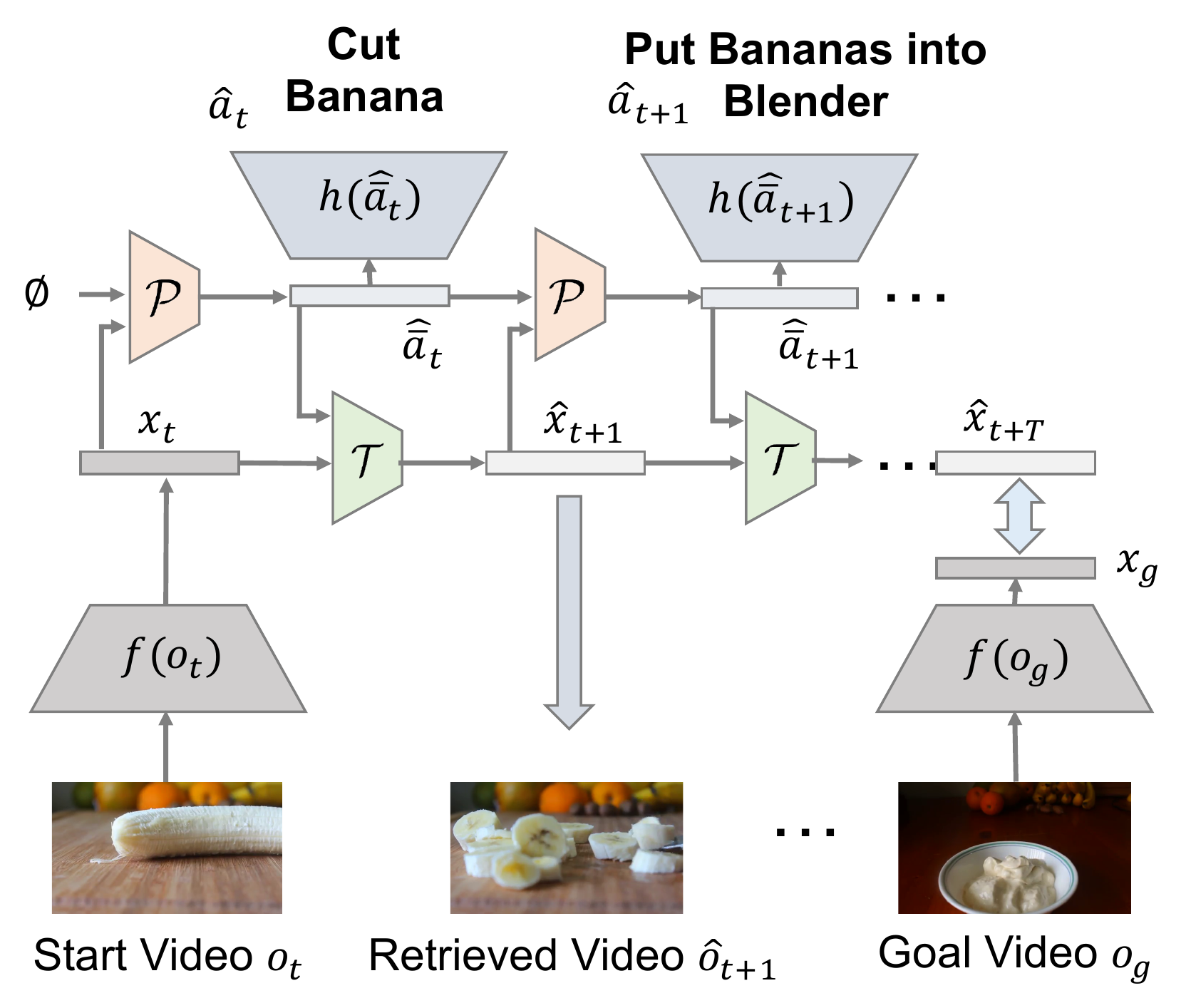}
    \vspace{-3mm}
  \caption{At inference time, our full model rollouts by sampling the action from the conjugate dynamics $\mathcal{P}$ and the next state from the forward dynamics $\mathcal{T}$. $\mathcal{P}$ essentially captures the applicable actions and improves planning performance.}
  \label{fig:inference}
    \vspace{-6mm}

\end{minipage}%
\hfill
\begin{minipage}{.49\textwidth}
  \centering
  \includegraphics[width=.9\linewidth]{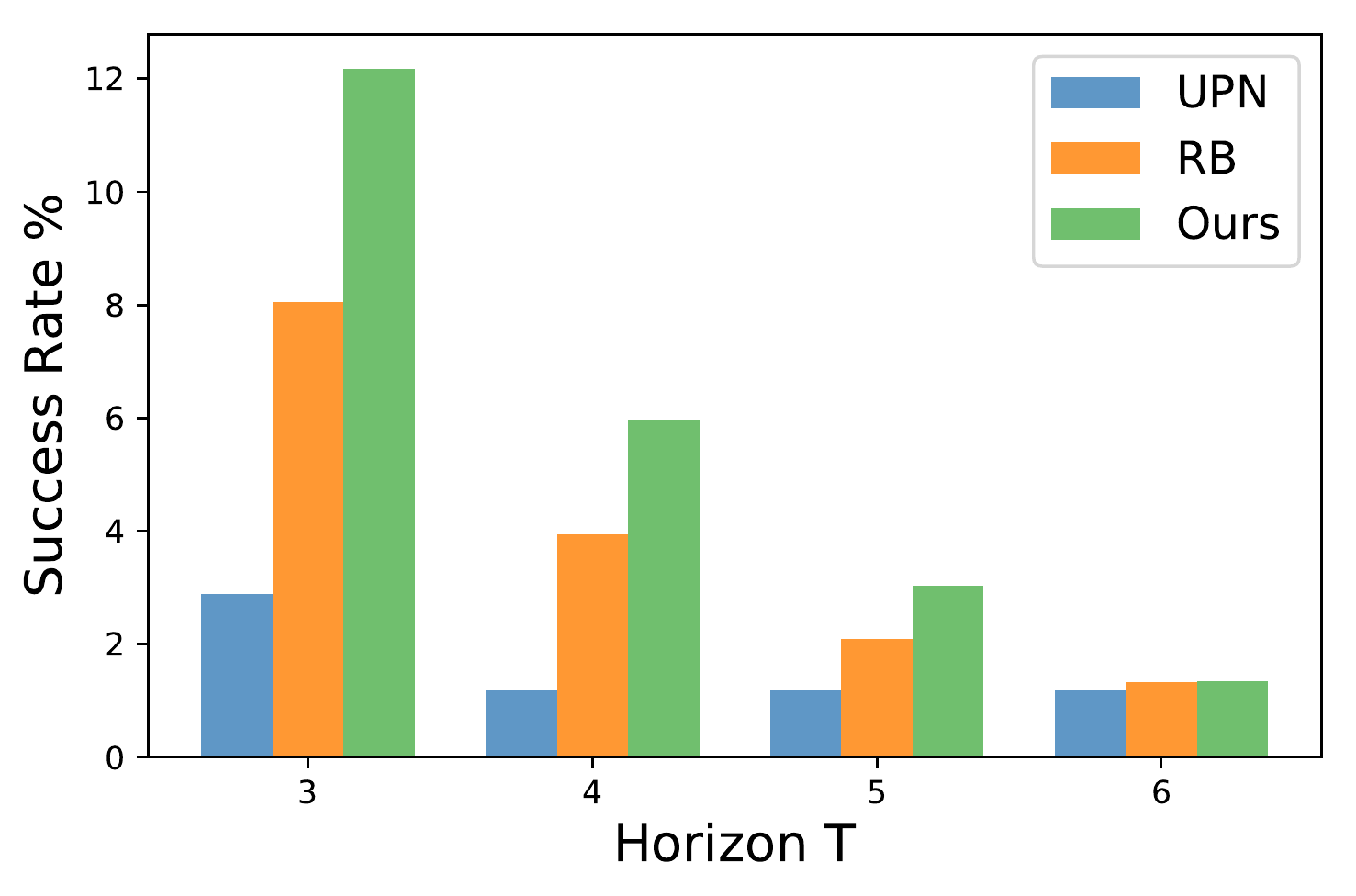}
    \vspace{-3mm}

  \caption{Our model consistently outperforms other baselines as the horizon of planning $T$ increases. When $T \ge 6$, the start observation and the visual goal are separated by 6 or more than 6 high-level actions. In this case the success rates of all baselines are less than $1\%$, suggesting there are more variations in the ways to reach the goal. }
  \label{fig:barplot}
    \vspace{-6mm}

\end{minipage}
\end{figure}

In this section, we discuss how to use DDN for planning in instructional videos. The general paradigm is illustrated in Figure~\ref{fig:inference}. At inference time, our full model rollouts by sampling the action from the conjugate dynamics $\mathcal{P}$ and the next state from the forward dynamics $\mathcal{T}$. $\mathcal{P}$ captures the applicable actions from the current state and improves planning performance. In the following, we first discuss how we can leverage the learned models in \secref{learn} to perform procedure planning in latent space. We then describe how our models can be applied to walkthrough planning~\cite{kurutach2018learning}, where the objective is to output the visual waypoints/subgoals between the current observation and the goal observation.

\begin{algorithm}[t]
    \caption{Procedure Planning}
    \label{alg:plan}
    \begin{algorithmic}[1]
    \State \textbf{Inputs:} Current and goal observations $o_t$, $o_g$; learned models $f(\cdot)$, $h(\cdot)$, $\mathcal{T}$, $\mathcal{P}$;  max iteration $\beta$, threshold $\epsilon$, beam size $\eta$
        \State $x \gets f(o_t)$, $x_g \gets f(o_g)$ 
        \State $x^* \gets x$, $\bar a \gets \emptyset$
        \State $q \gets PriorityQueue()$, $q \gets q \cup \{(x, \bar a)\}$
        \While {iteration $ < \beta$ and $||x^* - x_g||^2_2 > \epsilon$}
            \State $(x, \bar a) \gets Pop(q)$
            \State $\bar a' \sim \mathcal{P}(\cdot|x, \bar a)$ 
            \For {$\bar a_i \in \bar a'$}
                \State $x \gets \mathcal{T}(\cdot|x, \bar a_i)$ 
                \State $q \gets q \cup \{(x,\bar a_i)\}$
                \If {$||x - x_g||^2_2 < ||x^* - x_g||^2_2$}
                \State $x^* \gets x$ 
            \EndIf
            \EndFor
            \State $q \gets Sort(q)$ 
            \State $q \gets q[:\eta]$
        \EndWhile
        \State $\{\bar a_i^*\} \gets Backtracking(q, x^*)$
    \State \Return $\{h(\bar a_i^*)\}$
    \end{algorithmic}
\end{algorithm}

\noindent\textbf{Procedure Planning.} Using the learned models, we perform sampling-based forward planning~\cite{ghallab2004automated} to plan a sequence of actions to achieve the goal. The process is shown in Algorithm~\ref{alg:plan}. Given the current and the goal observations $o_t$ and $o_g$, we first map them to the latent space with $f(\cdot)$: $x_t = f(o_t)$, $x_g = f(o_g)$. In contrast to symbolic planning, we do not have a list of \emph{applicable} actions to apply in the search process. One additional advantage of having jointly learned the conjugate dynamics model $\mathcal{P}$ is that we can efficiently sample the actions to apply using $\mathcal{P}(\cdot|x_t, \bar a_{t-1})$. Based on $\bar a_t$ and $x_t$, we can obtain $x_{t+1}$ using $\mathcal{T}(\cdot|x, \bar a)$. 
The search process continues for a max iteration of $\beta$ and threshold $\epsilon$ while maintaining a priority queue of size $\eta$.

\noindent\textbf{Walkthrough Planning.} Kurutach \etal proposed walkthrough planning in ~\cite{kurutach2018learning}. 
The outputs of walkthrough planning can serve as visual signals of the subgoals to guide task executions. In addition, it is also helpful for interpretation by visualizing what the model has learned. The details of the process are shown in Algorithm~\ref{alg:walkthrough}. Given the pool of visual observations $\{o_i\}$, we can first construct the score matrix $R_{i,j}$ to capture the transition probability between two video clips $o_i$ and $o_j$ using our learned model $\mathcal{T}$ and $\mathcal{P}$. We can then perform walkthrough planning by finding the path of length $T$ that starts at $o_t$ and ends at $o_g$, while maximizing the total score. If the pool of video clips is all the clips in the same instructional video, then the problem is equivalent to finding a permutation function $b: \{1,2,...,T\} \rightarrow \{1,2,...,T\}$ that maximizes the total score along the permutation path, under the constraints that $b(1) = 1, b(T) = T$. 

\begin{algorithm}[t]
    \caption{Walkthrough Planning}
    \label{alg:walkthrough}
    \begin{algorithmic}[1]
    \State \textbf{Inputs:} All observations $\{o_i\}|_{i=1}^T$, learned models $f(\cdot)$, $g(\cdot)$, $\mathcal{T}$, $\mathcal{P}$, horizon $T$
        \State $b \gets \emptyset$
        \For {$i$ in \{$1\dots T$\}}
            \State $x_i \gets f(o_i)$
        \EndFor
        \For {$i$ in \{$1\dots T$\}}
            \For {$j$ in \{$1 \dots T$\}}
                \State $R_{i, j} \gets \sum_{a} \mathcal{T}(x_j|x_i,g(a)) \mathcal{P}(g(a)|x_i) $
            \EndFor
        \EndFor
    \State $\{o_{b(i)}\} \gets \arg \max_{b \in Perm(T)} \sum_{i=1}^T R_{b(i), b(i+1)}$
    \State \Return $\{o_{b(i)}\}$
    \end{algorithmic}
\end{algorithm}

\section{Experiments}
\label{sec:experiments}

We aim to answer the following questions in our experiments: 
(i) Can we learn plannable representations from real-world instructional videos? How does DDN compare to existing latent space planning methods? 
(ii) How important are the forward and conjugate dynamics? 
(iii) Can we apply DDN to walkthrough planning problem~\cite{kurutach2018learning} and how does it compare to existing methods? 

We answer the first two questions with ablation studies on challenging real-world instructional videos. In addition, we show how our model can generalize to various start and goal observations. For the last question, we directly apply our model to walkthrough planning~\cite{kurutach2018learning} and show that our approach outperforms existing approaches.

\noindent\textbf{Dataset.} To train and test our proposed model, we curate a video dataset of size $N$ that takes the form
\begin{align*}
\{(o_t^i, a_t^i, ..., a_{t+T-1}^i, o_{t+T}^i)\}_{i=1}^N. 
\end{align*}
Each example contains $T$ high-level actions, and each action can last from tens of seconds to 6 minutes. Our dataset is adapted from the 2750 labeled videos in CrossTask~\cite{zhukov2019cross}, averaging 4.57 minutes in duration, for a total of 212 hours. Each video depicts one of the 18 long-horizon tasks like {\it Grill Steak}, {\it Make Pancakes}, or {\it Change a Tire}. The videos have manually annotated temporal segmentation boundaries and action labels. Please note that since our method requires full supervision, we will leave how to utilize unlabeled instructional videos~\cite{zhukov2019cross,miech2019howto100m} as future work.  We encourage the reader refer to the supplementary material for data curation details. We divide the dataset into 70\%/30\% splits for training and testing. We use the precomputed features provided in CrossTask in all baselines and ablations for fair comparison. Each 3200-dimensional feature vector is a concatenation of the I3D, Resnet-152, and audio VGG features. The action space in CrossTask is given by enumerating all 105 combinations of the predicates and objects, which is shared across all 18 tasks. The distance function $\phi$ in CrossTask is the cross-entropy. Please note that our method does not assume specific forms of action space nor distance functions.

\noindent\textbf{Implementation Details.} The state mapping $f$ and action embedding $g$ are 128-dimensional. $\mathcal{T}$ is a 2-layer MLPs with 128 units that takes the outputs of $f$ and $g$ as inputs. 
The network $\mathcal{P}$ shares the state and the action embedding with $\mathcal{T}$ and we introduce recurrence to $\mathcal{P}$ by replacing the MLPs with a two-layer RNN of 128 hidden size, and concatenate the goal embedding as input. The action decoder $h$ takes the outputs of $\mathcal{P}$ as inputs and outputs actions as defined by the dataset. We set hyperparameters $\alpha = 0.001$, $\beta = 20T$, $\eta=20$, and $\epsilon = 1e-5$. We use a 5-fold cross validation on the training split to set the hyper-parameters, and report performance on the test set. We train our model for 200 epochs with batch size of 256 on a single GTX 1080 Ti GPU. We use Adam optimizer with learning rate of $1e-4$ and schedule the learning rate to decay with a decay factor $0.5$ and a patience epoch of $5$.

\subsection{Evaluating Procedure Planning} 

\begin{table*}[t]
\centering
\caption{Procedure planning results. Our model significantly outperforms baselines. With $\sim$10\% improvement of accuracy, our model is able to improve success rate by 8 times compared to \textit{Ours w/o $\mathcal{T}$}. This shows the importance of reasoning over the full sequence. * indicates re-implementations.}

 \label{tab:planning}
 \begin{tabular}{l|ccc|ccc}
\toprule
          & \multicolumn{3}{c|}{$T=3$}                     & \multicolumn{3}{c}{$T=4$}                     \\
\hline
          & Success Rate          & Accuracy          & mIoU           & Success Rate          & Accuracy          & mIoU           \\ \hline
Random    & \textless 0.01\%                 & 0.94\%              & 1.66\%                 & \textless 0.01\% & 0.83\% & 1.66\%              \\
RB~\cite{sun2019videobert}*         & 8.05\% & 23.3\% & 32.06\% & 3.95\% & 22.22\% & 36.97\%\\
WLTDO~\cite{ehsani2018let}*        &1.87\% &21.64\%&  31.70\%& 0.77\% &17.92 \%&26.43\%
\\
UAAA~\cite{farha2019uncertainty}*         & 2.15\%                 &  20.21\%              & 30.87\%                 & 0.98\%              & 19.86\% & 27.09\%\\
UPN~\cite{srinivas2018universal}*         & 2.89\%                 & 24.39\%              & 31.56\%                 & 1.19\%              & 21.59\% & 27.85\%\\

 \hline
Ours w/o $\mathcal{P}$ & \textless 0.01\%                 & 2.61\%              & 0.86\%                 & \textless 0.01\%             & 2.51\% & 1.14\% \\
Ours w/o $\mathcal{T}$ & 1.55\%                 & 18.66\%              & 28.81\%                & 0.65\%            & 15.97\% & 26.54\%  \\
Ours       & \textbf{12.18\%}                 & \textbf{31.29\%}              & \textbf{47.48\%}                 & \textbf{5.97\%}        & \textbf{27.10}\% & \textbf{48.46}\%     \\
\bottomrule

\end{tabular}
\end{table*}

In procedure planning, the inputs to the model are the start video clip $o_t$ and the goal video clip $o_g$, and the algorithm should output a sequence of $T$ actions that brings $o_t$ to $o_g$.

\noindent\textbf{Baselines.} We evaluate the capacity of our model in learning plannable representations from real-world videos and compare with the following baselines and ablations:

{\noindent \it  - Random Policy.} This baseline randomly selects an action from all actions. We include this baseline to show the empirical lower bound of performance.

{\noindent \it - Retrieval-Based (RB).} The procedure planning problem is formulated as a goal-conditioned decision making process. In contrast, one might approach this problem from a more static view which is analogous to~\cite{sun2019videobert}: Given the unseen start and goal visual observations, the retrieval-based baseline finds the nearest neighbor of the start and goal pair in the training set by minimizing the distance in the learned feature space, and then directly output the associated ground truth action labels. We use the same features as in our full model for fair comparison. 

{\noindent \it - WLTDO~\cite{ehsani2018let}.} Ehsani \etal  proposed an action planning model for egocentric dog videos. This baseline is a recurrent model for the planning task. Given two non-consecutive observations, it predicts a sequence of actions taking the state from the first observation to the second observation. We modify the author's implementation\footnote{https://github.com/ehsanik/dogTorch} to use the same features as our full model, and add softmax layer to output discrete actions.

{\noindent \it - Uncertainty-Aware Anticipation of Activities (UAAA)~\cite{farha2019uncertainty}.} This baseline has a two-step approach that uses RNN-HMM to infer the action labels in the observed frames, and then use an autoregressive model to predict the future action labels. We re-implement the model and modify it to condition on both the observed frames (start observations) and the visual goal. 

{\noindent \it  - Universal Planning Networks (UPN)~\cite{srinivas2018universal}.} UPN is the closest to ours among existing works. Similar to our approach, UPN also aims to learn a plannable representation using supervision from the imitation loss function at training. However, it assumes a continuous and differentiable action space to enable gradient-based planning, which might not be applicable to the discrete action space in CrossTask. We re-implement UPN and adapt it to output discrete actions by adding a softmax layer. 

{\noindent \it  - Ours w/o $\mathcal{T}$.} We compare to the ablation of our model without learning the forward dynamics $\mathcal{T}$, where we only $\mathcal{P}$ directly outputs the actions based on the previous action and the current state. We implement $\mathcal{P}$ with an RNN and concatenate the goal and start as input. In this case, this ablation is equivalent to a goal-conditional RNN policy directly trained with expert actions. This ablation can also be seen as a re-implementation of the RNN-based model in~\cite{abu2018will}.

{\noindent \it  - Ours w/o $\mathcal{P}$.} We compare to the ablation without learning the conjugate dynamics. In this case, the joint optimization of the forward dynamics $\mathcal{T}$ and the mapping $f(\cdot)$ to the latent space can easily overfit to the training sequences.

\noindent\textbf{Metrics.} We use three metrics for comparison. The first is \emph{success rate}. Although we do not have access to the underlying environment to evaluate the policies by executing them in simulation, we consider a plan as a success if all the actions in the plan are the same as those in the ground truth. This is a reasonable approximation because we consider a fixed number of steps, and there is less variation in the ways to complete the task. The second metric we consider is the \emph{accuracy} of the actions at each step, which does not require the whole sequence to match the ground truth as in the success rate metric, but only looks at each individual time step. We take the mean over the actions to balance the effect of repeating actions. The third metric we use is \emph{mean Intersection over Union} (mIoU), which is the least strict of all the metrics we use. We compare the IoU by $\frac{|\{a_t\} \cap \{a^*_t\}|}{|\{a_t\} \cup \{a_t^*\}|}$, where $\{a^*_t\}$ is the set of ground truth actions, and $\{a_t\}$ is the set of predicted actions. We use IoU to capture the cases where the model understands what steps are required, but fails to discern the order of actions.

\begin{figure*}[t]
  \centering
  \includegraphics[width=1.0\linewidth]{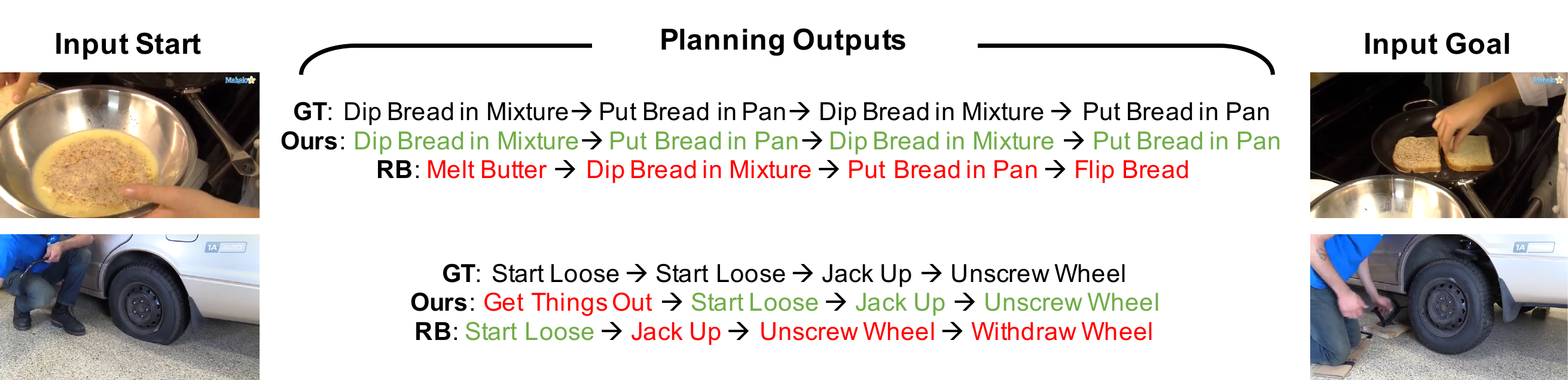}
    \vspace{-6mm}

  \caption{Examples of planned action sequences by DDN (ours) and the RB baseline. In the second example, DDN is not able to capture the subtle visual cues that the tool is already in the man’s hand, so there is no need to get the tools out.  }
  \label{fig:eccv_pp}
    \vspace{-6mm}

\end{figure*}

\begin{figure*}[t]
  \centering
  \includegraphics[width=1.0\linewidth]{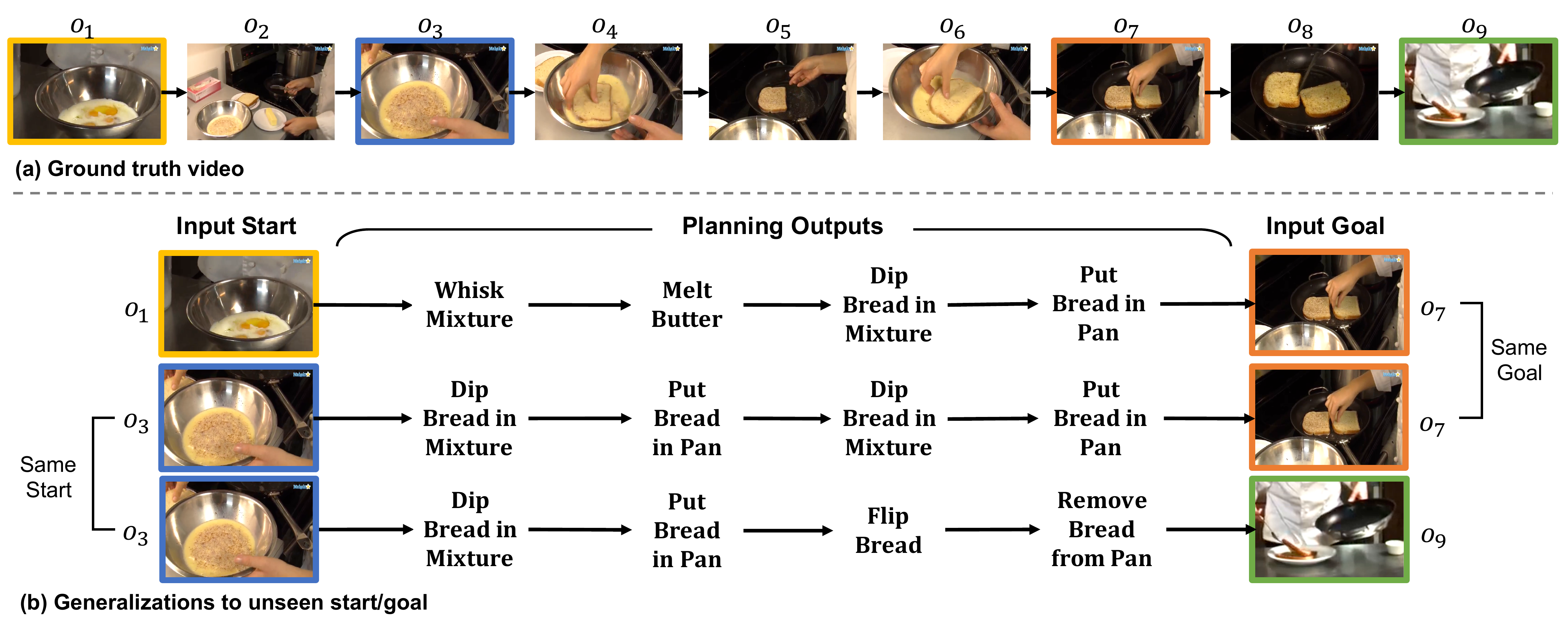}
  \vspace{-6mm}
  \caption{Examples of generalizations to unseen start/goal. (a) The 9 snapshot observations in making French toasts. (b) Our proposed model is robust to changes of start and goal observations among different stages in the video, and is able to output reasonable plans to achieve the goal and reason about the essential steps. For example, in the third row, the start $o_3$ and goal $o_9$ are separated by 6 actions. While the model has never seen such data during training, it successfully plans the 4 essential actions by merging repeated actions and further adding the pivotal actions of flipping and plating.}
  \label{fig:qual_pp}
  \vspace{-6mm}
\end{figure*}

\noindent\textbf{Results.} The results are shown in \tabref{planning}. UPN is able to learn representations that perform reasonably well compared to the random baseline. However, as the action space in instructional videos is not continuous, the gradient-based planner is not able to work well. Both WLTDO and UAAA perform similar to Ours w/o $\mathcal{T}$, which can be seen an RNN goal-conditional policy directly trained with imitation objectives. 
We also note that Ours w/o $\mathcal{P}$ cannot learn reasonable plannable representations without the conjugate dynamics. Our full model combines the strengths of planning and action imitation objective as conjugate constraints, which enables us to learn plannable representations from real-world videos to outperform all the baseline approaches on all metrics. In~\figref{barplot}, we further show our model's performance as the planning horizon increases. Our model consistently outperforms the RB baseline for all metrics because DDN explicitly imposes the structured priors using $\mathcal{T}$ and $\mathcal{P}$ to find the sequence of actions that reaches the goal. 

~\figref{eccv_pp} shows some qualitative results. We note that it is difficult for the predicted sequence of actions to be exactly the same as the ground-truth, which explains why the success rates in ~\tabref{planning} and ~\figref{barplot} are low in absolute value. The results are nevertheless semantically reasonable. Furthermore, as shown in~\figref{qual_pp}, it allows our model to generalize to different start observations when fixing the goal and vice versa.  
~\figref{qual_pp}(a) shows the 9 snapshot observations in making French toast. In the second row of~\figref{qual_pp}(b), we pick $o_3$ and $o_7$ as the start and goal observations and ask the model to plan for 4 actions to reach the goal. The model successfully recognizes that there are two French toasts and dip and pan fry the bread twice. In the third row, we change the goal to $o_9$. In this case, the start and goal are separated by 6 actions. While the model has never seen such data during training, it successfully plans the 4 essential actions by merging repeated actions and further adding the pivotal actions of flipping and plating. Similarly, in the first row, the model can also generalize to different start observations when fixing the goal.

\begin{table}[t]
    \centering
    \setlength\tabcolsep{2pt}
    \caption{Results for walkthrough planning. Our model significantly outperforms the baseline by explicitly reasoning what actions need to be performed first, and is less distracted by the visual appearances. * indicates re-implementations.}
    \label{tab:walkthrough}
\begin{tabular}{l|cc|cc}
\toprule
          & \multicolumn{2}{c|}{$T=3$}                     & \multicolumn{2}{c}{$T=4$}                     \\
\hline
          & Hamming          & Pair  Acc.          & Hamming           & Pair  Acc.          \\ \hline
Random    & 1.06                 & 46.85\%              & 1.95                 & 52.23\%              \\
RB~\cite{sun2019videobert}*         & 0.88 & 56.23\% & 1.80 & 55.42\% \\
VO~\cite{zeng2017visual}*         & 1.02                 & 49.06\%              & 1.99                 & 50.31\%              \\
Causal InfoGAN~\cite{kurutach2018learning}      & 0.57                 & 71.55\%              & 1.36                 & 68.41\%              \\ \hline
Ours w/o $\mathcal{T}$ & 0.99                 & 50.45\%              & 2.01                 & 47.39\%              \\
Ours w/o $\mathcal{P}$ & 0.33                 & 83.33\%              & 1.08                 & 77.11\%              \\
Ours       & \textbf{0.26}                 & \textbf{86.81\%}              & \textbf{0.88}                 & \textbf{81.21\%}             \\
\bottomrule

\end{tabular}

\end{table}

\subsection{Evaluating Intermediate States with Walkthrough Planning}

Given the start and goal video clips, we have shown that our model is able to plan a sequence of actions that brings the start to the goal. At the same time, our model also predicts a sequence of intermediate states. In this section, we evaluate and visualize these predicted intermediate states. Specifically, we show how our proposed method can be apply to \emph{walkthrough planning}~\cite{kurutach2018learning}, where the objective is to output the visual waypoints between the start and goal observations.

\noindent\textbf{Experimental Setup.} 
We evaluate the predicted intermediate state representations by using them to retrieve visual subgoals for task completion. 
In this way, the model only needs to predict lower-dimensional representations that can be used to retrieve the correct video clips from a pool of candidates. Specifically, we use all the video clips in the original video as the video clip pool. In this case, the task is equivalent to sorting the intermediate video clips while fixing the first and the last video clips.

\vspace{1mm}
\noindent\textbf{Additional Baselines.} We further compare to the following approaches for walkthrough planning:

{\noindent \it  - Visual Ordering (VO).} As the task is reduced to sorting a pool of video clips~\cite{zeng2017visual}, one baseline is to directly learn a model $V(o_1, o_2)$ to see if $o_1$ and $o_2$ are consecutive video clips in the same video.  Given $V(o_1, o_2)$, we can find the order of the candidate video clips by maximizing the total score given by $V$ using a greedy-based search. We learned $V(o_1, o_2)$ using the same setup as our $\mathcal{T}$, only that the actions $a$ are not used as the input.

{\noindent \it  - Causal InfoGAN (CIGAN)~\cite{kurutach2018learning}.} CIGAN learns plannable representations by maximizing the mutual information between the representations and the visual observations. Additionally, the latent space is assumed to follow the forward dynamics of a certain class of actions. CIGAN is able to perform walkthrough planning using minimal supervision. We modify the author's implementation\footnote{https://github.com/thanard/causal-infogan} to use the same features as our full model for fair comparison.

\noindent\textbf{Metrics.} 
Let $Y = (y_1, ..., y_T)$ be a sequence of the ground truth order, and $b: \{1,2,...,T\} \rightarrow \{1,2,...,T\}$ be the permutation function such that the prediction is $\hat{Y} = (y_{b(1)}, \dots, y_{b(T)})$. We use the following two metrics to evaluate the walkthrough planning outputs order:

{\noindent \it  - Hamming Distance:} counts the number of $\{i | i \neq b(i)\}$.

{\noindent \it  - Pairwise Accuracy:} calculates if the order between a pair $i$ and $j$ is respected by $b(i)$ and $b(j)$. It is given by $\frac{2}{T(T-1)} {\sum_{i<j, i \neq j}^{T} \{b(i) < b(j)\}}$.

\begin{figure}[t]
  \centering
  \includegraphics[width=0.99\linewidth]{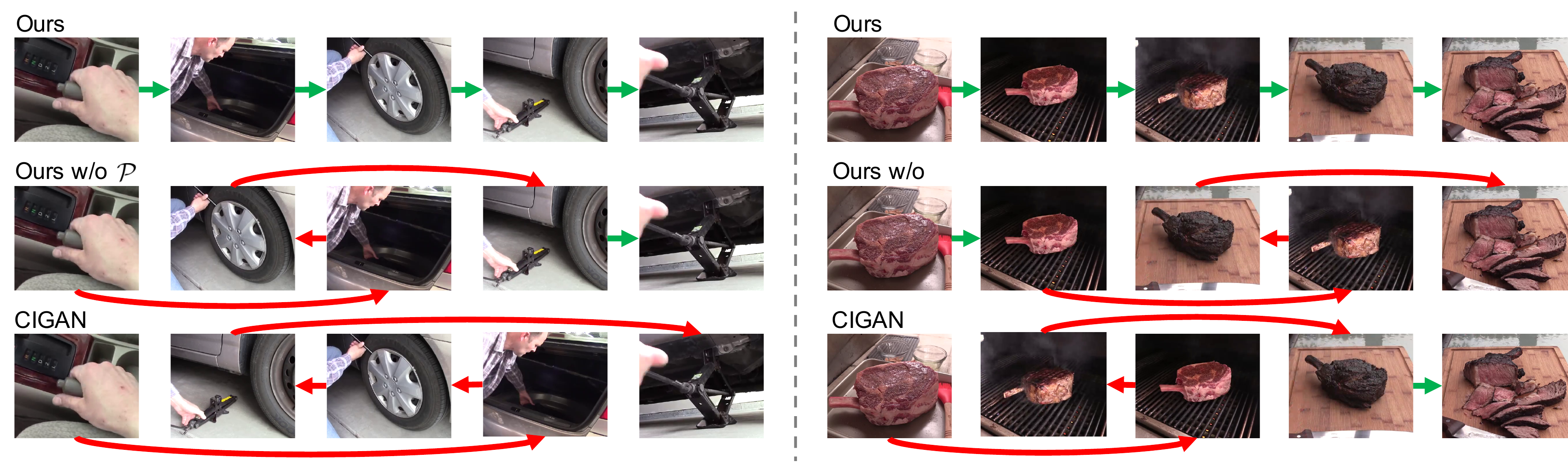}
  \vspace{-3mm}

  \caption{Examples of visualized predicted intermediate states by DDN (ours) and two baselines. Green/red arrows indicate correct/incorrect transitions between two visual steps. Our model successfully predicts sequence of state representations with correct ordering, while the baselines are more or less distracted by the visual appearances.}
    \vspace{-3mm}
  \label{fig:qualitative}
\end{figure}

\vspace{1mm}
\noindent\textbf{Results.} The results are shown in \tabref{walkthrough}. VO is unable to improve much over chance without modeling the actions. RB also struggles to perform well because it cannot handle the subsequence level variations that are not seen in the training data. By maximizing the mutual information with an adversarial loss, CIGAN\cite{kurutach2018learning} is able to learn reasonable models beyond Random without using action supervision. 
However, the complexity of the instructional videos requires explicit modeling of the forward dynamics conditioned on the semantic actions. Our full model learned for procedure planning successfully transfers to walkthrough planning and significantly outperforms all baselines on all metrics. It is interesting to see that Ours w/o $\mathcal{P}$ actually outperforms Ours w/o $\mathcal{T}$ in this case. Ours w/o $\mathcal{P}$ learns the forward dynamics $\mathcal{T}$ to directly anticipate the visual effect of an action $a$, which is better suited for walkthrough planning. By just using $\mathcal{P}$ the model is more accurate at predicting the actions, suggesting a trade-off between action and state space modeling. Our full model successfully combines the strength of the two and imposes the structured priors through conjugate constraints to both procedure planning and walkthrough planning. 

In Figure~\ref{fig:qualitative}, we visualize some examples of the predicted intermediates states, where the task is to change tires (left) and grill a steak (right). 
Our full model is able to predict a sequence of intermediate states with correct ordering. Specifically, the most challenging step in changing tire (left) is the second step, where the person goes to take the tools, and the video is visually different from the rest of the steps. Neither of the baseline models is able to understand that to perform the rest of the steps, the person needs to get the tools first.

\section{Conclusion}
We presented Dual Dynamics Networks (DDN), a framework for procedure planning in real-world instructional videos. DDN is able to learn plannable representations directly from unstructured videos by explicitly leveraging the structured priors imposed by the conjugate relationships between states and actions on the latent space. Our experimental results show our framework significantly outperforms a variety of baselines across different metrics. 
Our work can be seen as a step towards the goal of enabling autonomous agents to learn from real-world demonstrations and plan for complex tasks like humans.
In future work, we intend to incorporate object-oriented models to further explore the objects and predicates relations in visually complex environments.

\clearpage
%
%
\bibliographystyle{splncs04}
\bibliography{egbib}
\end{document}